\pdfoutput=1
\documentclass{article}
\usepackage[letterpaper,top=2cm,bottom=2cm,left=3cm,right=3cm,marginparwidth=1.75cm]{geometry}

\usepackage[utf8]{inputenc}
\usepackage[T1]{fontenc}
\usepackage{hyperref}[]
\usepackage{url}
\usepackage{booktabs}
\usepackage{natbib}
\usepackage{amsfonts}
\usepackage{nicefrac}
\usepackage{microtype}
\usepackage{xspace}
\usepackage{xcolor}
\usepackage{amsmath}
\usepackage{amssymb}
\usepackage[section]{placeins}
\usepackage{graphicx}
\usepackage{natbib}
\usepackage{booktabs}
\usepackage{xcolor}
\usepackage{colortbl} 
\usepackage{soul} 
\usepackage{tabularx}
\usepackage{stfloats}
\usepackage{algorithm,algorithmicx}
\usepackage[noend]{algpseudocode}
\usepackage{algcompatible}
\graphicspath{{./figures/}}
\usepackage{multirow}
\usepackage[most]{tcolorbox}
\usepackage{mathtools}
\usepackage{shortvrb}
\usepackage{subcaption}
\usepackage{pifont}

\usepackage{amsmath,amsfonts,bm}

\def\eqref#1{equation~\ref{#1}}

\def\1{\bm{1}}

\DeclareMathAlphabet{\mathsfit}{\encodingdefault}{\sfdefault}{m}{sl}
\SetMathAlphabet{\mathsfit}{bold}{\encodingdefault}{\sfdefault}{bx}{n}

\usepackage{enumitem}

\usepackage{amssymb} 
\usepackage{pifont}  

\newcommand{\cmark}{\ding{51}}%
\newcommand{\xmark}{\ding{55}}%

\definecolor{lightgrey}{HTML}{dcdbdb}
\definecolor{lightblue}{HTML}{E8F0FE}
\definecolor{gray}{HTML}{9aa0a6}
\definecolor{lightpink}{HTML}{F48FB1}
\definecolor{lightred}{HTML}{FFCBC9}
\definecolor{lightcyan}{HTML}{80DEEA}
\newcommand{\cc}[0]{\cellcolor{lightblue}}
\newcommand{\cg}[0]{\cellcolor{lightgrey}}
\newtcolorbox{mybox}[2][]
  {colback = black!5!white, colframe = black!75!black, fonttitle = \bfseries,
    colbacktitle = black!100!black, enhanced, 
    attach boxed title to top left={yshift=-2.2mm,xshift=4mm},
    title=#2,#1}

\usepackage{soul}

\setcitestyle{round}

\newcommand{\github}{\raisebox{-1.5pt}{\includegraphics[height=1.05em]{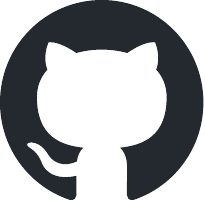}}\xspace}
\newcommand{\huggingface}{\raisebox{-1.5pt}{\includegraphics[height=1.05em]{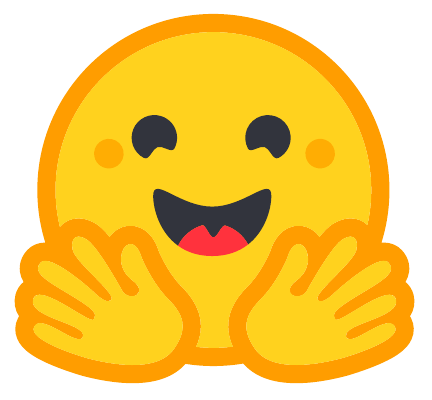}}\xspace}

\title{Dream-VL \& Dream-VLA: Open Vision-Language and Vision-Language-Action Models with Diffusion Language Model Backbone}

\author{%
  \textbf{Jiacheng Ye}$^{1,*}$ \quad
  \textbf{Shansan Gong}$^{1,*}$ \quad 
  \textbf{Jiahui Gao}$^{2,\dag,\diamondsuit}$ \quad
  \textbf{Junming Fan}$^2$ \and
  \textbf{Shuang Wu}$^2$ \quad
  \textbf{Wei Bi}$^\ddag$ \quad
  \textbf{Haoli Bai}$^2$ \quad
  \textbf{Lifeng Shang}$^2$ \quad
  \textbf{Lingpeng Kong}$^{1,\dag}$ \\
  \vspace{0.5em}
  $^1$The University of Hong Kong \quad 
  $^2$Huawei Technologies \quad 
\\
 \huggingface\href{https://huggingface.co/Dream-org/Dream-VL-7B}{\textbf{Dream-VL}} \quad  \huggingface\href{https://huggingface.co/Dream-org/Dream-VLA-7B}{\textbf{Dream-VLA}} \quad  \github\href{https://github.com/DreamLM/Dream-VLX}{\textbf{DreamLM/Dream-VLX}} 
}

\date{December 23, 2025}

\begin{document}

\maketitle
\begin{abstract}
While autoregressive Large Vision-Language Models (VLMs) have achieved remarkable success, their sequential generation often limits their efficacy in complex visual planning and dynamic robotic control.
In this work, we investigate the potential of constructing Vision-Language Models upon diffusion-based large language models (dLLMs) to overcome these limitations.
We introduce \textbf{Dream-VL}, an open diffusion-based VLM (dVLM) that achieves state-of-the-art performance among previous dVLMs. Dream-VL is comparable to top-tier AR-based VLMs trained on open data on various benchmarks but exhibits superior potential when applied to visual planning tasks. 
Building upon Dream-VL, we introduce \textbf{Dream-VLA}, a dLLM-based Vision-Language-Action model (dVLA) developed through continuous pre-training on open robotic datasets.
We demonstrate that the natively bidirectional nature of this diffusion backbone serves as a superior foundation for VLA tasks, inherently suited for action chunking and parallel generation, leading to significantly faster convergence in downstream fine-tuning.
Dream-VLA achieves top-tier performance of 97.2\% average success rate on LIBERO, 71.4\% overall average on SimplerEnv–Bridge, and 60.5\% overall average on SimplerEnv–Fractal, surpassing leading models such as $\pi_0$ and GR00T-N1.
We also validate that dVLMs surpass AR baselines on downstream tasks across different training objectives. We release both Dream-VL and Dream-VLA to facilitate further research in the community. 
\end{abstract}

\let\thefootnote\relax
\footnotetext[1]{$^{*}$Equal contribution $^{\dag}$Corresponding author $^{\diamondsuit}$Project Lead $^{\ddag}$Independent Researcher }
\let\thefootnote\arabic

\section{Introduction}
\label{sec:intro}
Vision-language models are reshaping AI across diverse domains -- from medical diagnosis and scientific discovery to autonomous systems and robotic manipulation~\citep{tu2024towards,zhang2024generalist,zhang2024comprehensive,zhou2024vision,palme,rt2}. As these applications grow in complexity, a fundamental capability emerges as critical: visual planning -- the ability to perceive visual context, reason about goals, and formulate coherent action sequences. Whether analyzing multi-step experimental procedures in scientific papers, guiding a surgical robot through delicate operations, orchestrating a delivery drone's route, or enabling a home robot to organize a kitchen, success hinges on the capacity for sophisticated long-horizon planning.

Large vision-language models (VLMs) have demonstrated remarkable progress in grounding language with visual representations, powering applications from multimodal dialogue to scene understanding. Building on these foundations, vision-language-action (VLA) models extend these capabilities to action generation, enabling agents to interact with physical and virtual environments. However, current VL and VLA models overwhelmingly rely on autoregressive large language models (LLMs) as their backbone. This architectural choice presents a fundamental bottleneck: the autoregressive paradigm, trained on next-token prediction, struggles with tasks requiring long-horizon planning and global reasoning~\citep{bachmann2025pitfallsnexttokenprediction,ye2024beyond,weng2025tracefutureprobabilisticreasoning}. Besides, its sequential generation nature suffers from error accumulation during inference, further hindering global reasoning capabilities~\citep{zhang2023language,rohatgi2025}.

Diffusion large language models (dLLMs) offer a compelling alternative to address these challenges~\citep{inceptionlabs_mercury,nie2025llada,dream2025,geminidiffusion}. Unlike autoregressive models that generate tokens sequentially, diffusion models learn to iteratively refine noisy sequences into coherent outputs. This iterative refinement process naturally encourages global coherence, making dLLMs particularly well-suited for planning tasks where long-range dependencies and goal-oriented reasoning are critical~\citep{dream2025, zhang2025exploringpowerdiffusionlarge}. Moreover, the parallel decoding capability of diffusion language models offers computational advantages during inference~\citep{inceptionlabs_mercury,geminidiffusion,wang2025diffusionllmsfasterthanarinference,wu2025fast}. Given these promising characteristics in language modeling, a natural question arises: can these advantages translate to vision-language tasks, where planning and coherent reasoning are equally critical?

To explore this potential, we build \textbf{Dream-VL}, a diffusion LLM-based vision-language model (dVLM) for general visual understanding, and \textbf{Dream-VLA}, a diffusion LLM-based vision-language-action model (dVLA) for robotic manipulation. We conduct comprehensive experiments across visual understanding, visual planning, and embodied action benchmarks.
Our results show that:
\begin{itemize}
    \item Dream-VL achieves comparable performance to top-tier autoregressive VLMs trained on open data, and significantly surpasses existing dVLMs such as LLaDA-V~\citep{you2025lladav} and Dimple~\citep{yu2025dimple}. Notably, Dream-VL shows superior performance over strong AR baselines on visual planning tasks (e.g.,  LIBERO~\citep{liu2023libero}, ViPlan~\citep{merler2025viplan}) that require long-horizon planning. 
    \item Built upon Dream-VL, Dream-VLA establishes top-tier performance on both the evaluated SimplerEnv~\citep{simplerenv} and LIBERO~\citep{liu2023libero} benchmark, surpassing top VLA models such as $\pi_0$~\citep{black2024pi0} and GR00T-N1~\citep{gr00t}. Notably, Dream-VLA consistently outperforms the strong AR-based baseline OpenVLA-OFT~\citep{openvlaoft} across diverse finetuning objectives, demonstrating that a natively bidirectional dVLM can serve as a superior backbone for VLA tasks.
\end{itemize}
We attribute these strong results to three inherent advantages of diffusion-based VLMs over autoregressive models: \textbf{(1)} Bidirectional attention enables richer information fusion between visual and text features; \textbf{(2)} The text planning capability in dLLMs enhances visual planning in VLMs by generating global plans aligned with predefined goals; \textbf{(3)} Diffusion VLMs naturally support action chunking and parallel generation without architectural modification, which are crucial for low-level action prediction in VLA models and enable faster convergence during downstream fine-tuning.

\begin{figure}[t]
    \centering
    \includegraphics[width=\textwidth]{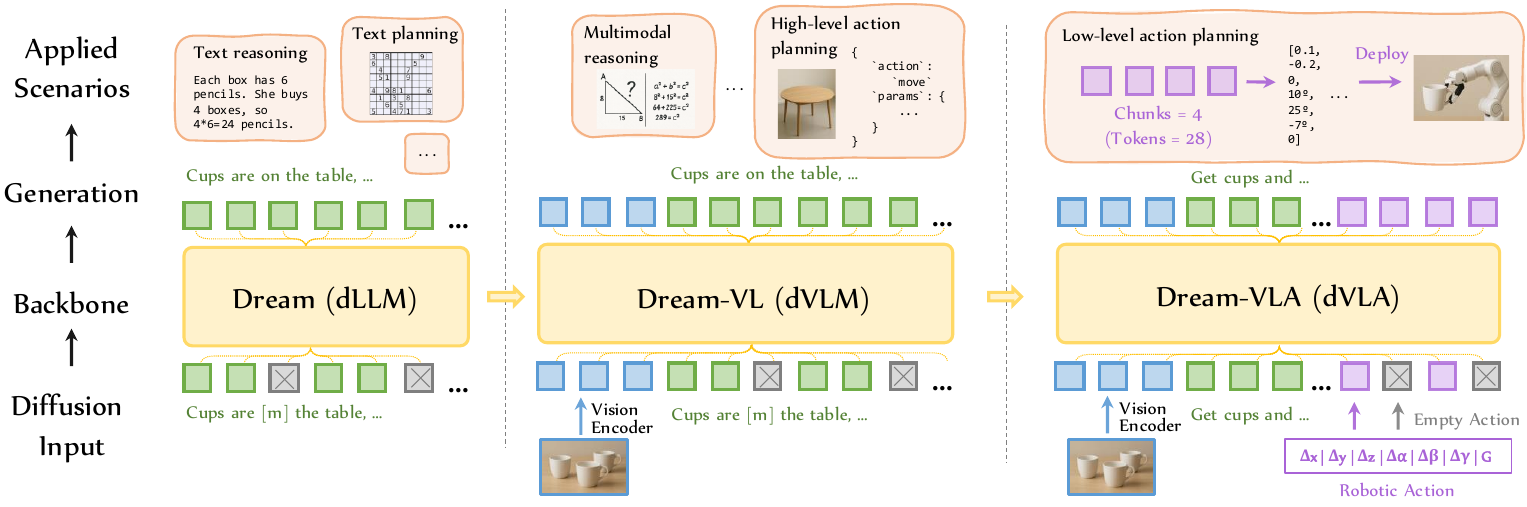} 
    \caption{Overview of the Dream family, with bi-directional masked diffusion modeling. Built on the diffusion language model Dream-7B, we introduce Dream-VL, a state-of-the-art diffusion VLM that demonstrates strong multimodal understanding, reasoning, and effective long-horizon planning. Building further on Dream-VL, we perform VLA pretraining to obtain the first pretrained diffusion-based VLA model (dVLA), which serves as a strong backbone for downstream VLA tasks.}
    \label{fig:overview}
\end{figure}
\section{Related Work}
\label{sec:related}
\paragraph{Diffusion Language Model}
Diffusion models are now considered a promising alternative to traditional autoregressive generation methods, as they support flexible generation orders and work through iterative refinement. The idea of using discrete diffusion for language modeling was first explored by \citet{austin2021structured, hoogeboom2021argmax}, which provided an initial framework for text modeling. Since then, various studies have been proposed to improve text diffusion models~\citep{campbell2022continuous,li2022diffusion,chen2023analog,gong2022diffuseq,zheng2023reparameterized,sahoo2024simple,shi2025simplifiedgeneralizedmaskeddiffusion,ou2024your,xu2025energy,liu2025think,zhang2025target,zheng2025continuously,pynadath2025candi,havasi2025edit,yang2025powerful,Dreamon2025,kang2025ladir,campbell2025self}. Larger models have been trained from scratch, such as SEDD~\citep{lou2023discrete}, Plaid~\citep{gulrajani2023likelihood}, MDGA~\citep{ni2025diffusion} and  LLaDA~\citep{nie2025llada,zhu2025llada,zhu2025lladamoe}. 
Meanwhile, methods that adapt pretrained autoregressive models for diffusion, such as DiffuLLaMA~\citep{gong2025scalingdiffusionlanguagemodels} and Dream~\citep{dream2025}, have led to significant reductions in the training costs of dLLMs.
DLLMs have exhibited promising capabilities in planning~\citep{zhang2023planner,ye2024diffusion,ye2024beyond,ye2025implicit} and code generation~\citep{khanna2025mercury,gong2025diffucoder,xie2025dream}. Recently, \citet{geminidiffusion,arriola2025blockdiffusion,ma2025dkv,wang2025diffusionllmsfasterthanarinference,cheng2025sdar,wu2025fast, wu2025fastdllmv2,gat2025set,liu2025tidar} have further advanced dLLMs by improving generation quality and inference efficiency.

\paragraph{Vision-Language Model (VLM)}
Autoregressive vision-language models have achieved remarkable success across various multimodal tasks. Pioneering works~\citep{flamingo,li2023blip,llava,instructblip,llavaimproved,liu2024llavanext,chen2024expanding,chen2024internvl,nvlm,wang2024qwen2,openai2024gpt4ocard, bai2025qwen2,guo2025mammothvl} have demonstrated strong capabilities in visual understanding, reasoning, and instruction following by aligning pretrained vision encoders with large language models. 
Extending dLLMs to vision-language tasks remains an emerging research area. 
LaViDA~\citep{lilavida} explores a discrete diffusion transformer with multi-view image encoding and masked-denoising objectives. LLaDA-V~\citep{you2025lladav} applies masked diffusion with visual instruction tuning, which supports parallel decoding and controllable infilling. MMaDA~\citep{yang2025mmada} proposes a unified diffusion transformer that aims to align reasoning between text and vision through chain-of-thought supervision and reinforcement learning. Dimple~\citep{yu2025dimple} addresses the training instability of discrete diffusion by employing a hybrid autoregressive-then-diffusion paradigm. 
Lavida-O~\citep{li2025lavidao} introduces a unified masked diffusion model for multimodal understanding and generation, incorporates planning and iterative self-reflection in image generation and editing tasks. LLaDA-MedV~\citep{dong2025llada-med} utilizes masked diffusion in medical applications, integrating imaging and textual data for diagnosis.

\paragraph{Vision-Language-Action Model (VLA)}
Building a generalist robotic policy has long been a central goal in embodied AI. Driven by the rapid advances in autoregressive Vision-Language, researchers have increasingly focused on Vision-Language-Action (VLA) models to leverage strong multimodal generalization for robotic control~\citep{palme,rt2,pai0,openvla,cogact,wen2025tinyvla,team2025gemini,gr00t}. 
Pioneering work such as RT-2~\citep{rt2} established the VLA paradigm by jointly fine-tuning VLMs on web-scale data and low-level robot demonstrations. OpenVLA~\citep{openvla} is further introduced as the first open-source VLA model.
To address the computational demands of these models, subsequent work has introduced token compression~\citep{pifast} and action chunking~\citep{zhao2023learning,openvlaoft}, enabling higher-frequency control and faster adaptation.
Build upon diffusion models~\citep{diffusion_policy,pearce2022imitating,reuss2023goal} for action prediction, a VLM with an additional action expert architecture has also been actively explored to predict coherent continuous actions~\citep{cogact,black2024pi0,wen2025tinyvla,gr00t,shukor2025smolvla,zhang2025dreamvla}.
While standard VLAs rely on autoregressive modeling, discrete diffusion VLA has also been recently studied for robotic control~\citep{discretediffusion,lladavla} and autonomous driving~\citep{cui2025vilad,li2025discrete,ma2025dvlm,xu2025wam}.
DiscreteDiffusionVLA~\citep{discretediffusion} and LLaDA-VLA~\citep{lladavla} have introduced discrete masked diffusion as an alternative by directly finetuning specific robotic tasks on an autoregressive dLLM backbone~\citep{discretediffusion} or dVLM backbone~\citep{lladavla} without robotic pretraining.

\section{Dream-VL:  Diffusion-based VLM Built on Dream 7B}

To fully and controllably analyze the model characteristics, we first train Dream-VL, a well-aligned diffusion-based VLM built upon Dream 7B~\citep{dream2025}.

\subsection{Setup}
\paragraph{Multi-modal Training.}

Following MAmmoTH-VL~\citep{guo2025mammothvl}, we collect 12M open-source multimodal data to train the VL model, which contains instruction-response pairs from diverse, real-world tasks such as mathematical problem-solving, OCR, and domain-specific reasoning. We incorporate a Qwen2ViT~\citep{wang2024qwen2} module to encode visual inputs into latent features, which are concatenated with text features as model input. The model is trained using the same discrete diffusion loss as Dream 7B, followed by a multi-stage training paradigm. The detailed training parameters of each stage are shown in Table~\ref{tab:vl-params}.
\begin{table}[ht]
\caption{Training parameters of Dream-VL.}
\scalebox{0.86}{
\begin{tabular}{lccc}
\toprule
\textbf{} & Stage-1 & Stage-2 & Stage-3 \\
\hline
Dataset & LCS & Single Image (SI) & Single, Multi-Image \& Video \\
\#Samples & 558K & 10M & 2M \\
Vision Tower & Qwen2ViT & Qwen2ViT & Qwen2ViT \\
LLM Backbone & Dream-v0-Instruct-7B & Dream-v0-Instruct-7B & Dream-v0-Instruct-7B \\
Trainable Model Parameters & Projector: 25.7M & Full Model: 8.3B & Full Model: 8.3B \\
Batch Size & 512 & 256 & 256 \\
Model Max Length & 8192 & 8192 & 8192 \\
Learning Rate / Epoch & 1e-3 / 1ep & 1e-5 / 1ep + 5e-6 / 2ep & 5e-6 / 1ep \\
\bottomrule
\end{tabular}}
\label{tab:vl-params}
\end{table}

\begin{table}[b!]
    \centering
    \caption{Performance comparison on multi-discipline knowledge and mathematical reasoning benchmarks. ``Open Data'' indicates whether the multimodal training data is open-source. Best results among diffusion models are \textbf{bolded}. SI indicates Single Image.} \label{tab:vl_sota_comparison1}
    \setlength{\tabcolsep}{3pt} 
    \resizebox{\textwidth}{!}{%
    \begin{tabular}{llc|ccccc|cc}
        \toprule
        & & \textbf{Open} & \multicolumn{5}{c|}{\textbf{Multi-Discipline Knowledge}} & \multicolumn{2}{c}{\textbf{Mathematical Reasoning}} \\
        \textbf{Model} & \textbf{LLM Backbone} & \textbf{Data} & \textbf{MMMU} & \textbf{MMMU Pro} & \textbf{MMStar} & \textbf{MMBench} & \textbf{SeedBench} & \textbf{MathVista} & \textbf{MathVerse} \\
        \midrule
        \multicolumn{10}{l}{\textit{Proprietary Models}} \\
        GPT-4o~\citep{openai2024gpt4ocard} & - & \xmark & 69.1 & 49.7 & 64.7 & 82.1 & 76.2 & 63.8 & 50.2 \\
        Gemini-1.5 Pro~\citep{Reid2024Gemini15} & - & \xmark & 65.8 & 44.4 & 59.1 & 73.9 & 76.0 & 63.9 & - \\
        Claude 3.5 Sonnet~\citep{claude35} & - & \xmark & 68.3 & 48.0 & 62.2 & 79.7 & 72.2 & 67.7 & - \\
        \midrule
        \multicolumn{10}{l}{\textit{Autoregressive VLMs}} \\
        DeepSeek-VL2~\citep{Wu2024DeepSeekVL2MV} & DeepSeek-MoE 27B & \xmark & 51.1 & - & 61.3 & 79.6 & - & 62.8 & - \\
        Qwen2.5-VL~\citep{Bai2025Qwen25VLTR} & Qwen2.5-7B & \xmark & 58.6 & - & 63.9 & 83.5 & - & 68.2 & 49.2 \\
        InternVL3~\citep{Zhu2025InternVL3EA} & Qwen2.5-7B & \xmark & 62.7 & - & 68.2 & 83.4 & - & 71.6 & 39.8 \\
        MiMo-VL-RL~\citep{Yue2025MiMoVLTR} & MiMo-7B & \xmark & 66.7 & 40.3 & - & 84.4 & - & 81.5 & 71.5 \\
        Cambrian-1~\citep{tong2024cambrian} & LLaMA3-8B & \cmark & 42.7 & 14.7 & - & 74.6 & 73.3 & 49.0 & - \\
        Llava-CoT-11B~\citep{xu2024llavacotletvisionlanguage} & LLaMA3.2-11B & \cmark & 48.9 & 18.5 & 57.6 & 75.0 & 75.2 & 54.8 & 24.2 \\
        Molmo-8B-D~\citep{deitke2024molmo} & Qwen2 7B & \cmark & 45.3 & 18.9 & 50.5 & 73.6 & 74.1 & 51.6 & 21.5 \\
        LLaVA-OV (SI)~\citep{li2024llava} & Qwen2-7B & \cmark & 47.3 & 16.8 & 60.9 & 80.5 & 74.8 & 56.1 & 26.9 \\
        LLaVA-OV~\citep{li2024llava} & Qwen2-7B & \cmark & 48.8 & 18.7 & 61.7 & 80.8 & 75.4 & 63.2 & 26.2 \\
        MAmmoTH-VL (SI)~\citep{guo2025mammothvl} & Qwen2.5-7B & \cmark & 49.4 & 26.0 & 55.4 & 83.0 & 73.3 & 67.6 & 35.0 \\
        MAmmoTH-VL~\citep{guo2025mammothvl} & Qwen2.5-7B & \cmark & 50.8 & 25.3 & 63.0 & 83.4 & 76.0 & 67.6 & 34.2 \\
        \midrule
        \multicolumn{10}{l}{\textit{Diffusion VLMs}} \\
        LLaDA-V~\citep{you2025lladav} & LLaDA 8B & \cmark & 48.6 & 18.6 & 60.1 & - & 74.8 & 50.6 & 28.5 \\
        MMaDA~\citep{yang2025mmada} & LLaDA 8B & \xmark & 30.2 & - & - & 68.5 & 64.2 & - & - \\
        FUDOKI~\citep{Wang2025FUDOKIDF} & Janus-1.5B & \cmark & 34.7 & - & - & 73.6 & 68.2 & - & - \\
        Dimple~\citep{yu2025dimple} & Dream 7B & \cmark & 45.2 & - & - & 74.6 & - & 42.4 & - \\
        LaViDa-D~\citep{lilavida} & Dream 7B & \cmark & 42.6 & - & - & 73.8 & - & 42.1 & 24.1 \\
        \cc \textbf{Dream-VL (SI)} &\cc  Dream 7B &\cc \cmark & \cc \textbf{51.8} & \cc 25.8 & \cc \textbf{60.2} & \cc \textbf{83.6} & \cc 75.9 & \cc \textbf{64.5} & \cc 30.7 \\
        \cc \textbf{Dream-VL} & \cc Dream 7B & \cc \cmark & \cc \textbf{52.2} &\cc  \textbf{26.0} & \cc 59.9 & \cc 83.0 & \cc \textbf{76.4} &\cc  63.1 & \cc \textbf{31.5} \\
        \bottomrule
    \end{tabular}%
    }
\end{table}
\begin{table}[t]
    \centering
    \caption{Performance comparison on chart \& document understanding and real-world multimodal interactions benchmarks. ``Open Data'' indicates whether the multimodal training data is open-source. Best results among diffusion models are \textbf{bolded}.} \label{tab:vl_sota_comparison2}
    \setlength{\tabcolsep}{3pt} 
    \resizebox{0.95\textwidth}{!}{%
    \begin{tabular}{llc|cccc|c}
        \toprule
        & & \textbf{Open} & \multicolumn{4}{c|}{\textbf{Chart \& Doc Understanding}} & \textbf{Interact} \\
        \textbf{Model} & \textbf{LLM Backbone} & \textbf{Data} & \textbf{AI2D} & \textbf{ChartQA} & \textbf{InfoVQA} & \textbf{DocVQA} & \textbf{RealWorldQA} \\
        \midrule
        \multicolumn{8}{l}{\textit{Proprietary Models}} \\
        GPT-4o~\citep{openai2024gpt4ocard} & - & \xmark & 94.2 & 85.7 & 79.2 & 92.8 & 76.5 \\
        Gemini-1.5 Pro~\citep{Reid2024Gemini15} & - & \xmark & 94.4 & 87.2 & 81.0 & 93.1 & 70.4 \\
        Claude 3.5 Sonnet~\citep{claude35} & - & \xmark & 94.7 & 90.8 & 49.7 & 95.2 & 60.1 \\
        \midrule
        \multicolumn{8}{l}{\textit{Autoregressive VLMs}} \\
        DeepSeek-VL2~\citep{Wu2024DeepSeekVL2MV} & DeepSeek-MoE 27B & \xmark & 81.4 & 86.0 & 61.3 & - & 68.4 \\
        Qwen2.5-VL~\citep{Bai2025Qwen25VLTR} & Qwen2.5-7B & \xmark & 83.9 & 87.3 & 82.6 & 95.7 & 68.5 \\
        InternVL3~\citep{Zhu2025InternVL3EA} & Qwen2.5-7B & \xmark & 85.2 & 86.6 & 76.8 & 92.7 & 70.8 \\
        MiMo-VL-RL~\citep{Yue2025MiMoVLTR} & MiMo-7B & \xmark & 83.5 & 91.7 & 88.0 & 95.7 & - \\
        Cambrian-1~\citep{tong2024cambrian} & LLaMA3-8B & \cmark & 73.0 & 73.3 & 41.6 & 77.8 & 64.2 \\
        Llava-CoT-11B~\citep{xu2024llavacotletvisionlanguage} & LLaMA-3.2-11B & \cmark & - & 67.0 & 44.8 & - & - \\
        Molmo-8B-D~\citep{deitke2024molmo} & Qwen2 7B & \cmark & 81.0 & 84.1 & 72.6 & 92.2 & 70.7 \\
        LLaVA-OV (SI)~\citep{li2024llava} & Qwen2-7B & \cmark & 81.6 & 78.8 & 65.3 & 86.9 & 65.5 \\
        LLaVA-OV~\citep{li2024llava} & Qwen2-7B & \cmark & 81.4 & 80.0 & 68.8 & 87.5 & 66.3 \\
        MAmmoTH-VL (SI)~\citep{guo2025mammothvl} & Qwen2.5-7B & \cmark & 83.4 & 85.9 & 74.8 & 93.8 & 71.3 \\
        MAmmoTH-VL~\citep{guo2025mammothvl} & Qwen2.5-7B & \cmark & 84.0 & 86.2 & 73.1 & 93.7 & 69.9 \\
        \midrule
        \multicolumn{8}{l}{\textit{Diffusion VLMs}} \\
        LLaDA-V~\citep{you2025lladav} & LLaDA 8B & \cmark & 77.8 & 78.3 & 66.3 & 83.9 & 63.2 \\
        Dimple~\citep{yu2025dimple} & Dream 7B & \cmark & 74.4 & 63.4 & - & - & - \\
        LaViDa-D~\citep{lilavida} & Dream 7B & \cmark & 69.0 & 61.0 & 36.2 & 56.1 & - \\
        \cc \textbf{Dream-VL (SI)} & \cc Dream 7B & \cc \cmark & \cc 80.6 & \cc \textbf{86.8} & \cc \textbf{81.4} & \cc \textbf{94.4} & \cc \textbf{68.4} \\
        \cc \textbf{Dream-VL} & \cc Dream 7B & \cc \cmark & \cc \textbf{81.2} & \cc 84.5 & \cc 81.0 & \cc \textbf{94.4} & \cc 66.3 \\
        \bottomrule
    \end{tabular}%
    }
\end{table}

\paragraph{Evaluation.}
Our experimental evaluation incorporates a wide range of vision-language understanding benchmarks: (1) \textbf{Multidisciplinary Knowledge} (MMStar~\citep{chen2024we}, MMMU~\citep{yue2023mmmu}, MMMU-Pro~\citep{yue2024mmmu}, SeedBench~\citep{li2023seedbenchbenchmarkingmultimodalllms}, MMBench~\citep{liu2024mmbenchmultimodalmodelallaround}, MMvet~\citep{yu2023mmvetevaluatinglargemultimodal}); (2) \textbf{Mathematical Reasoning} (Mathverse~\citep{zhang2024mathversedoesmultimodalllm}, Mathvista~\citep{lu2024mathvistaevaluatingmathematicalreasoning}); (3) \textbf{Chart/Doc Understanding} (AI2D~\citep{kembhavi2016diagram}, ChartQA~\citep{masry2022chartqa}, DocVQA~\citep{mathew2020docvqa}, InfoVQA~\citep{mathew2021infographicvqa}); (4) \textbf{Multimodal Interaction }(RealworldQA~\citep{grok15v}, WildVision~\citep{lu2024wildvisionevaluatingvisionlanguagemodels}, Llava-Wilder-Small~\citep{li2024llavanext-strong}); and (5) \textbf{Multi-image/Video Understanding }(MuirBench~\citep{wang2024muirbenchcomprehensivebenchmarkrobust}, SeedBench-Video~\citep{li2023seedbenchbenchmarkingmultimodalllms}, MLVU~\citep{zhou2024mlvucomprehensivebenchmarkmultitask}, VideoMME~\citep{fu2024video}).

\begin{table}[h!]
    \centering
    \caption{Performance comparison on multi-image and video understanding benchmarks. ``Open Data'' denotes open-source multimodal training data. Best results among diffusion models are \textbf{bolded}.}
    \label{tab:video_results}
    \setlength{\tabcolsep}{4pt}
    \resizebox{0.95\textwidth}{!}{%
    \begin{tabular}{llc|cccc}
        \toprule
        & & \textbf{Open} & \multicolumn{4}{c}{\textbf{Multi-image and Video Understanding}} \\
        \textbf{Model} & \textbf{LLM Backbone} & \textbf{Data} & \textbf{Seed-video} & \textbf{VideoMME} & \textbf{MuirBench} & \textbf{MLVU} \\
        \midrule
        \multicolumn{7}{l}{\textit{Proprietary Models}} \\
        GPT-4o~\citep{openai2024gpt4ocard} & - & \xmark & - & 71.9 & 68.0 & 64.6 \\
        Gemini-1.5 Pro~\citep{Reid2024Gemini15} & - & \xmark & - & - & - & - \\
        Claude 3.5 Sonnet~\citep{claude35} & - & \xmark & - & - & - & - \\
        \midrule
        \multicolumn{7}{l}{\textit{Autoregressive VLMs}} \\
        DeepSeek-VL2~\citep{Wu2024DeepSeekVL2MV} & DeepSeek-MoE 27B & \xmark & - & - & - & - \\
        Qwen2.5-VL~\citep{Bai2025Qwen25VLTR} & Qwen2.5-7B & \xmark & - & 65.1 & - & - \\
        InternVL3~\citep{Zhu2025InternVL3EA} & Qwen2.5-7B & \xmark & - & 66.3 & - & - \\
        MiMo-VL-RL~\citep{Yue2025MiMoVLTR} & MiMo-7B & \xmark & - & 67.4 & - & - \\
        Cambrian-1~\citep{tong2024cambrian} & LLaMA3-8B & \cmark & - & - & - & - \\
        Llava-CoT-11B~\citep{xu2024llavacotletvisionlanguage} & LLaMA-3.2-11B & \cmark & - & - & - & - \\
        Molmo-8B-D~\citep{deitke2024molmo} & Qwen2 7B & \cmark & - & - & - & - \\
        LLaVA-OV~\citep{li2024llava} & Qwen2-7B & \cmark & 56.9 & 58.2 & 41.8 & 64.7 \\
        MAmmoTH-VL~\citep{guo2025mammothvl} & Qwen2.5-7B & \cmark & 57.1 & 58.8 & 55.1 & 64.7 \\
        \midrule
        \multicolumn{7}{l}{\textit{Diffusion VLMs}} \\
        \cc LLaDA-V & \cc LLaDA 8B & \cc \cmark &\cc  53.7 & \cc 56.1 &\cc  48.3 & \cc 59.4 \\
        \cc \textbf{Dream-VL} &\cc  Dream 7B &\cc  \cmark &\cc  \textbf{58.1} &\cc  \textbf{61.5} & \cc \textbf{51.2} & \cc \textbf{61.1} \\
        \bottomrule
    \end{tabular}%
    }
\end{table}

\subsection{Experimental Results}

\paragraph{Dream-VL exhibits substantial improvements over existing diffusion-based VLMs.}
Tables~\ref{tab:vl_sota_comparison1}-\ref{tab:video_results} show the experimental results. Dream-VL significantly outperforms previous diffusion-based VLMs, demonstrating the effectiveness of our approach. Compared to LLaDA-V, Dream-VL achieves superior performance while using comparable training data ($\sim$12M in Dream-VL vs. $\sim$13M in LLaDA-V). This improvement can be attributed to the enhanced capabilities of the base Dream model and differences in training strategies. When compared to other Dream-based VLMs (e.g., Dimple/LaViDa), the performance gains primarily result from the substantially larger training dataset ($\sim$12M vs. $\sim$2M samples), highlighting the importance of sufficient training for diffusion-based VLMs to fully explore their potential.

\paragraph{Dream-VL achieves competitive performance with autoregressive VLMs and shows distinct advantages.} 
Compared to state-of-the-art autoregressive models trained on similar open datasets, Dream-VL achieves competitive overall performance, 
with notable advantages on multi-discipline visual reasoning (MMMU) and document understanding tasks (DocVQA, ChartQA), demonstrating its distinct advantages.
These results suggest that diffusion-based modeling can match or exceed autoregressive approaches in some scenarios. However, a performance gap remains when compared to leading closed-data autoregressive models (e.g., Qwen2.5-VL), indicating opportunities for further improvement through data scaling and model optimization.

\subsection{Analysis on Planning Ability}

\begin{figure}[t]
    \centering
    \includegraphics[width=\textwidth]{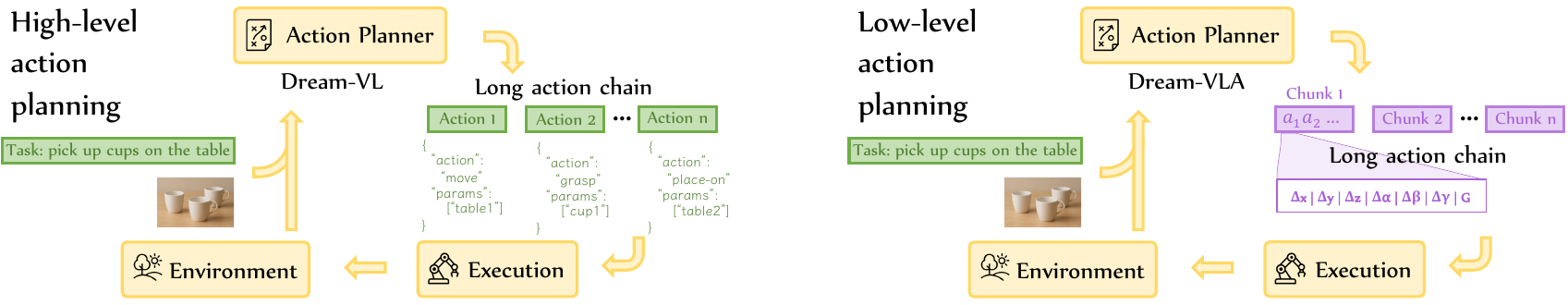} 
    \caption{High-level and low-level action planning. Both forms of planning require multi-turn interaction with the environment, typically modeled as a Markov Decision Process. The task requires the model to perform long-horizon planning with high consistency. High-level planning operates in a symbolic or semantic action space, producing abstract commands. Low-level planning directly focuses on precise robot behavior, handling fine-grained, continuous control such as grasping and motion trajectories. Additional robotic training is required for bridging the vision-language model with executable robotic actions.}
    \label{fig:action-plan}
\end{figure}

Most of the aforementioned tasks primarily reflect the model's general understanding capabilities. 
In this section, we investigate the effectiveness of dLLMs as VL backbones for \textbf{planning tasks}. Specifically, we focus on two types of planning: high-level action planning that operates in a symbolic or semantic action space, and low-level action planning that focuses on precise control of robot behavior. The two types of planning are shown in Figure~\ref{fig:action-plan}.

\subsubsection{Dream-VL for High-level Action Planning}

\paragraph{Task Formulation.}
High-level action planning requires generating a sequence of symbolic actions $\{a_1, a_2, \ldots, a_T\}$ that transform an initial scene state $s_0$ to a goal state $s_g$, given visual observations $I$ and a natural-language instruction $l$. Each action $a_t$ represents a structured, discrete operation (e.g., \texttt{pick(block1)}, \texttt{place(block2, left)}) executable by downstream controllers. This task evaluates whether a model can understand the scene, ground entities, and reason over multi-step procedures.

\begin{figure}[h]
\centering
\small
\begin{tabular}{|p{0.47\textwidth}|p{0.47\textwidth}|}
\hline
\multicolumn{1}{|c|}{\textbf{Grounding Mode}} & \multicolumn{1}{c|}{\textbf{Planning Mode}} \\
\hline
\begin{minipage}[t]{0.45\textwidth}
\vspace{4pt}
\textbf{Output Format:} The model outputs a binary ``yes'' or ``no'' indicating whether the state description is satisfied in the current scene. The grounding results assist the planner in producing executable actions.
\vspace{4pt}

\textbf{Example:} \\
\texttt{Input:} \texttt{[image]} + \textit{``Is the hardback on top of the shelf?''} \\[2pt]
\texttt{Output:} \texttt{"No"}
\vspace{4pt}
\end{minipage}
&
\begin{minipage}[t]{0.45\textwidth}
\vspace{4pt}
\textbf{Output Format:} The model outputs a JSON list of symbolic actions. For each iteration, only the first action is executed; the scene is updated and fed into the next iteration until the goal is satisfied or the maximum steps are reached.
\vspace{4pt}

\textbf{Example:} \\
\texttt{Input:} \texttt{[image]} + \textit{``Place the hardback on shelf.''} \\[2pt]
\texttt{Output:} \\
\texttt{\{"plan": [\{"action": "navigate-to",} \\
\texttt{\ \ \ "parameters": ["shelf\_1"]\},} \\
\texttt{\ \ \{"action": "place-on",} \\
\texttt{\ \ \ "parameters": ["hardback\_1"]\}]}
\texttt{\}}
\end{minipage} \\
\hline
\multicolumn{2}{|l|}{
\begin{minipage}[t]{0.93\textwidth}
\vspace{4pt}
\textbf{Domains:} (1) \textit{BlockWorlds} — colored blocks are arranged and manipulated under symbolic relational predicates (\texttt{on}, \texttt{clear}, \texttt{incolumn}); (2) \textit{Household} — everyday objects in a home environment must be manipulated and moved in more complex spatial and commonsense scenarios. \\
\end{minipage}
} \\
\hline
\end{tabular}
\caption{Output formats and domains of the ViPlan benchmark.}
\label{fig:viplan_output}
\end{figure}

\paragraph{Setup.} 
We evaluate Dream-VL on the ViPlan benchmark~\citep{merler2025viplan}. In each task instance, the input consists of an image $I$ depicting the scene and a language instruction $l$ describing the goal. The benchmark operates in two modes: (1) \textit{Grounding mode}, where the model outputs a binary response indicating whether a state description is satisfied in the current scene, and (2) \textit{Planning mode}, where the model generates a sequence of symbolic actions to achieve the specified goal. 
ViPlan comprises two domains: \textit{BlockWorlds}, involving colored blocks manipulated under symbolic relational predicates (e.g., \texttt{on}, \texttt{clear}, \texttt{incolumn}), and \textit{Household}, featuring everyday objects in complex spatial scenarios. Each domain includes three difficulty tiers (simple, medium, hard), varying in scene complexity, required plan length, and subtlety of visual cues.  Beyond grounding relations in the image, the model is also required to produce structured, symbolic actions that are similar to function or tool calls. Thus, ViPlan evaluates a hybrid capability that blends vision-textual understanding and planning, and precise symbolic planning. We report task success rate (task accuracy), measured by whether the final state satisfies the goal description, and action accuracy, defined as the proportion of valid actions among all generated actions.

\begin{figure}[t]
    \centering
    \begin{subfigure}[b]{0.49\textwidth}
        \centering
        \includegraphics[width=\textwidth]{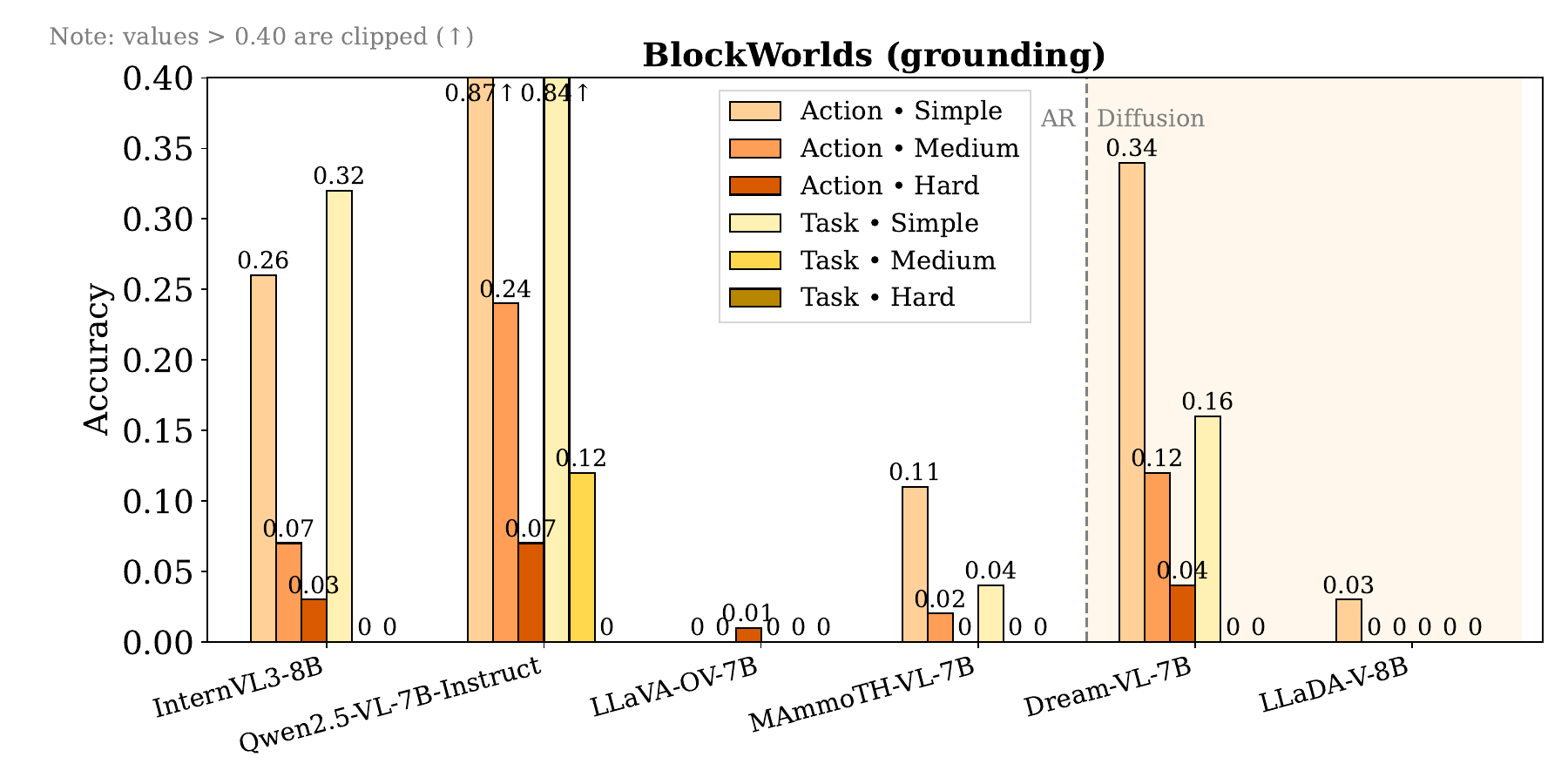}
        \label{fig:sub1}
    \end{subfigure}
    \hfill 
    \begin{subfigure}[b]{0.49\textwidth}
        \centering
        \includegraphics[width=\textwidth]{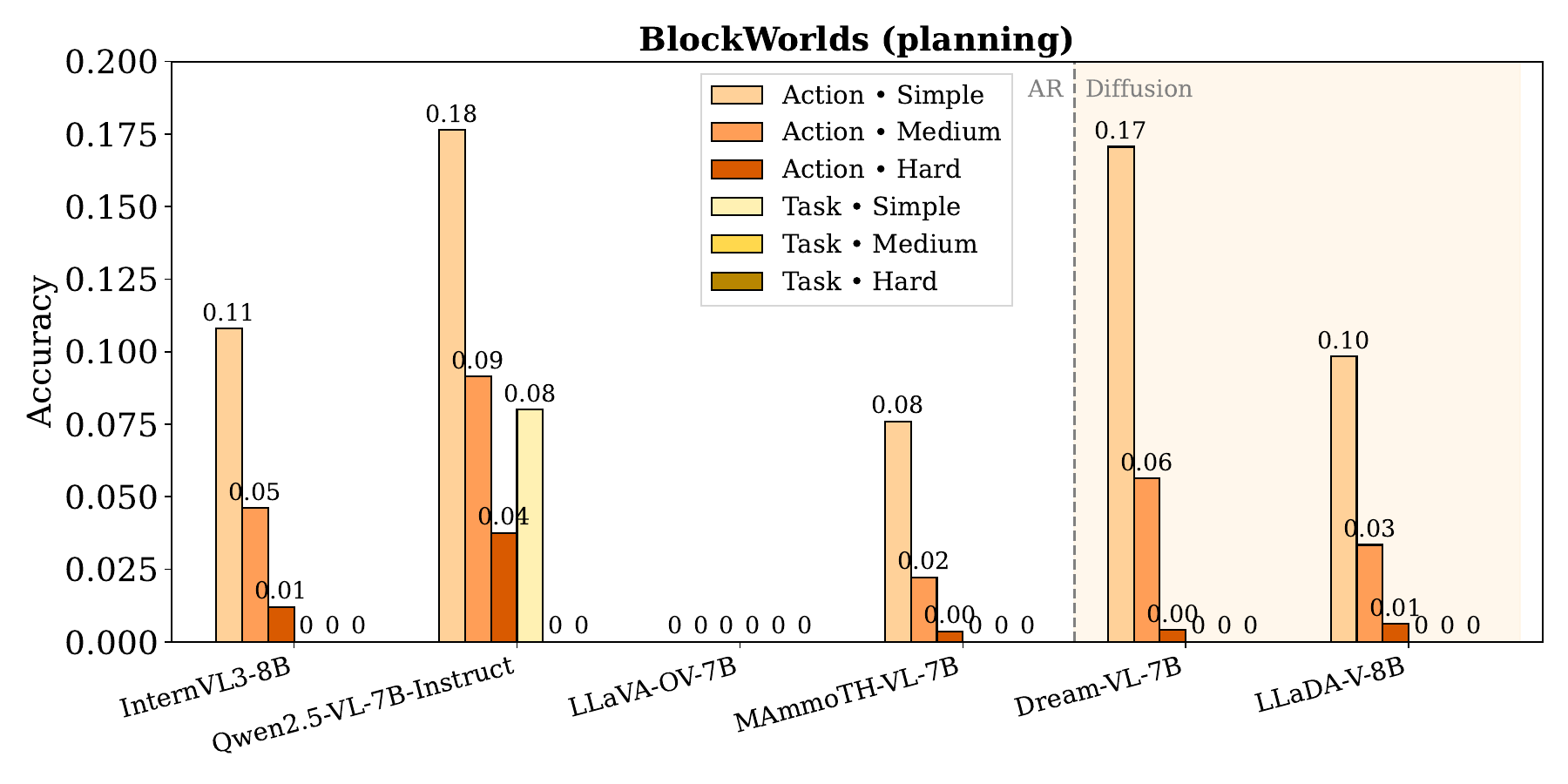}
        \label{fig:sub2}
    \end{subfigure}
    \begin{subfigure}[b]{0.49\textwidth}
        \centering
        \includegraphics[width=\textwidth]{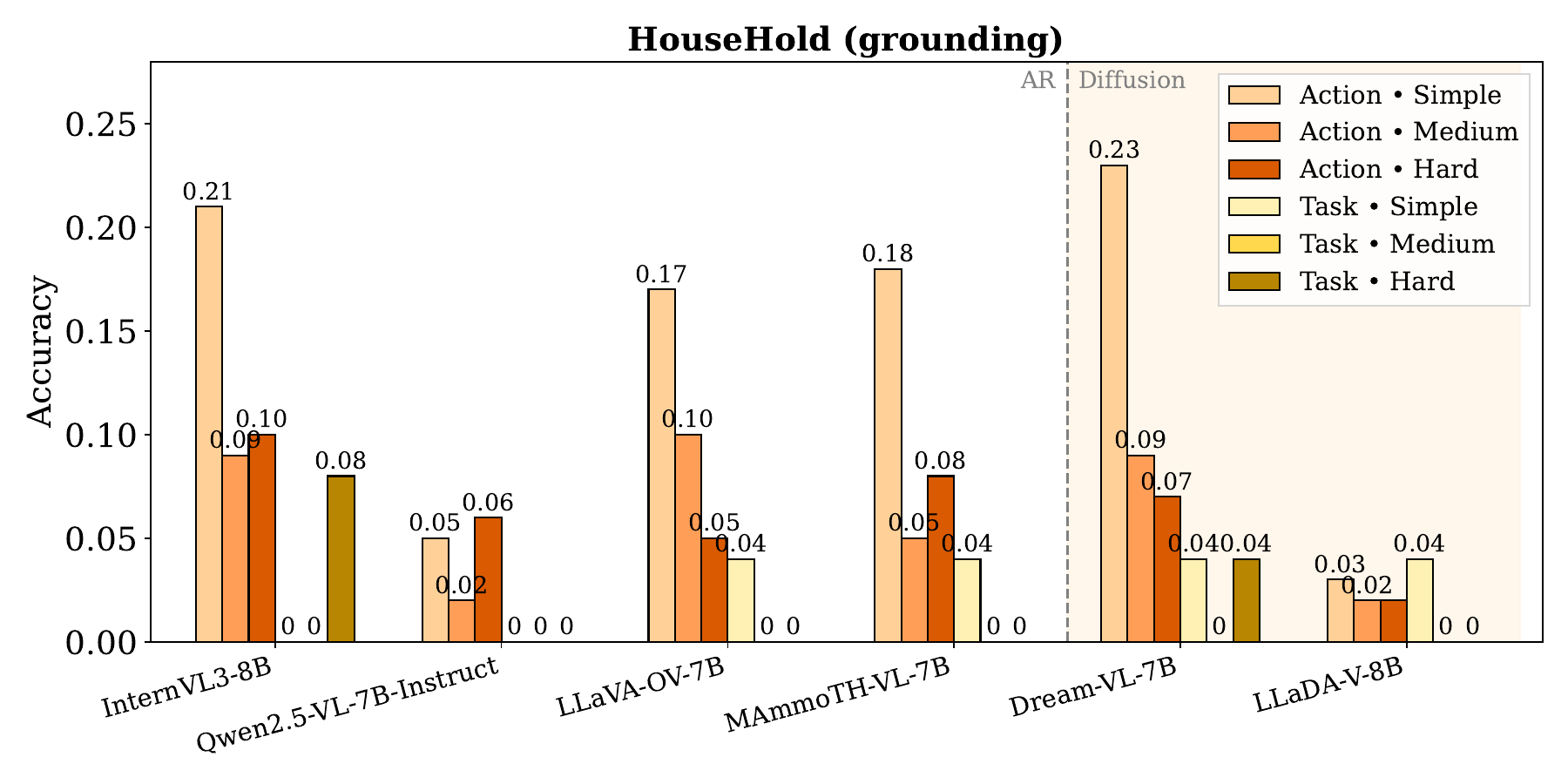}
        \label{fig:sub3}
    \end{subfigure}
    \hfill
    \begin{subfigure}[b]{0.49\textwidth}
        \centering
        \includegraphics[width=\textwidth]{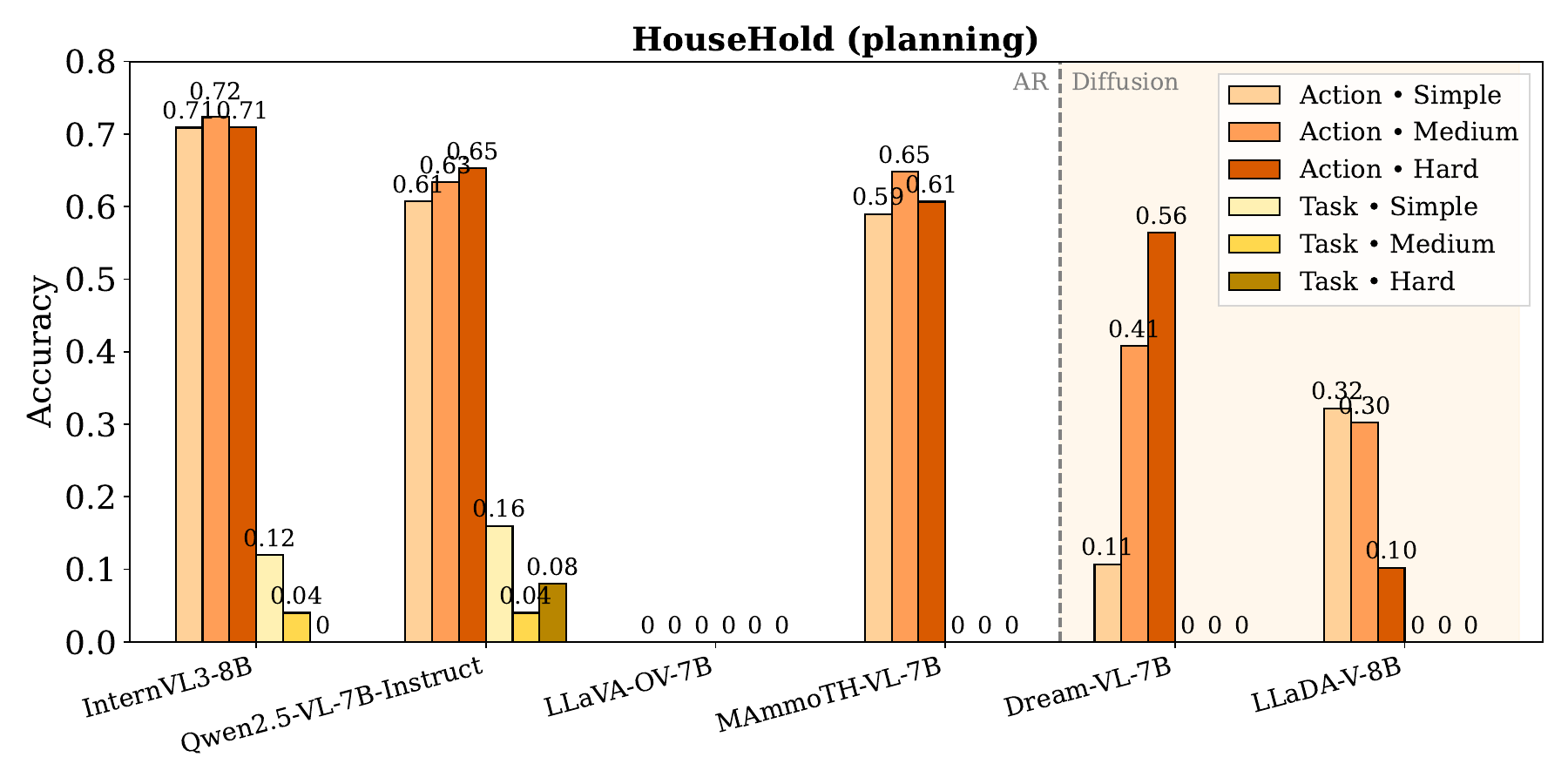}
        \label{fig:sub4}
    \end{subfigure}
    \vspace{-5mm}
    \caption{Performance comparison for VLMs on the ViPlan benchmark~\citep{merler2025viplan}.}
    \label{fig:viplan_bench}
\end{figure}

\paragraph{Results and Analysis.}
We evaluate Dream-VL on the ViPlan benchmark in a zero-shot manner, comparing it against both autoregressive and diffusion-based vision-language models. The results are presented in Figure~\ref{fig:viplan_bench}.  We have the following observations:

 \begin{itemize}
     \item  Compared to autoregressive VLMs, Dream-VL achieves competitive performance with InternVL3. While a performance gap remains relative to Qwen2.5-VL-Instruct, which has undergone extensive vision-language alignment, Dream-VL demonstrates promising capabilities as a diffusion-based alternative. More importantly, when controlling for training data and procedures, specifically comparing Dream-VL with MAmmoTH-VL-7B, which shares similar training recipes, Dream-VL achieves higher accuracy across most settings. This controlled comparison validates the benefits of the diffusion-based architecture for high-level planning tasks.

     \item Within the diffusion model family, Dream-VL exhibits substantial improvements and consistently outperforms LLaDA-V across all evaluation modes and difficulty tiers. We attribute this advantage to the strong planning capabilities inherited from the base Dream 7B model, which is good at text planning tasks such as Sudoku.
    
     \item Across both model families, Dream-VL narrows the performance gap with top autoregressive baselines, with particularly notable advantages in the Household grounding task. This suggests that diffusion-based modeling may offer specific benefits for complex visual grounding scenarios that require understanding spatial and commonsense relations in realistic environments.
    \end{itemize}

\begin{mybox}[colback=gray!10]{Highlights}
    \begin{itemize}
      \item Under controlled comparison, Dream-VL outperforms the autoregressive baseline (MAmmoTH-VL-7B) on most grounding and planning scenarios, highlighting the architectural advantages of diffusion-based modeling for high-level planning.
      \item Dream-VL achieves state-of-the-art performance among diffusion-based VLMs, consistently outperforming LLaDA-V across all evaluation modes.
    \end{itemize}
\end{mybox}

\subsubsection{Dream-VL for Low-level Action Planning}

\paragraph{Task Formulation.}

Low-level action planning requires predicting continuous motor controls that can be directly executed by controllers. Given visual observations $I$ and a natural-language instruction $l$, the model generates a sequence of robot control actions $\{a_1, a_2, \ldots, a_T\}$.  Following~\citet{rt2}, for each action, the model predicts 7-dimensional robot control to indicate the delta end-effector pose ($\Delta$pos, $\Delta$rotation, gripper).

\begin{figure}[h]
    \centering
    \includegraphics[width=\textwidth]{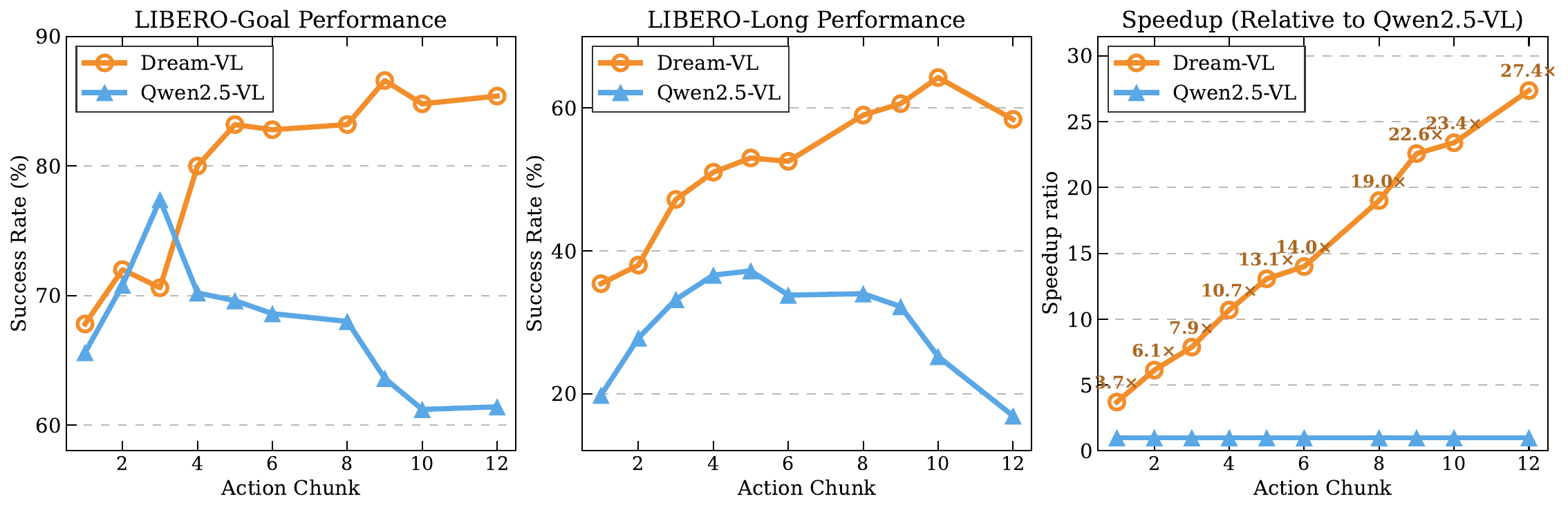}
    \caption{Comparison of model performance and speed with varying action chunk sizes.}
    \label{fig:low_level}
\end{figure}

\paragraph{Setup.}
We conducted experiments on the LIBERO~\citep{liu2023libero} benchmark. We use the LIBERO-Goal subtask, which tests procedural learning through varying task goals with fixed objects, and the LIBERO-Long subtask, which includes 10 long-horizon tasks with different objects, layouts, and goals.
To see the ability of Dream-VL in such low-level action planning, we first discretize each dimension of the robot actions separately into one of 256 bins following~\citep{rt2,openvla}. Then we finetune with autoregressive loss for Qwen2.5-VL and discrete diffusion loss for Dream-VL.
We follow the same preprocessed pipeline as in OpenVLA~\citep{openvla} and do not use wrist camera image and robot proprioceptive state in this experiment. Both Qwen2.5-VL and Dream-VL are directly finetuned on the LIBERO data without robotic pretraining. We set the training action chunk to be 8 to be consistent with OpenVLA, and compare the three models in Table~\ref{tab:libero_long_ablation}. The diffusion timestep is set to be 1 for Dream-VL. Additionally, we launch another comparison between Qwen2.5-VL and Dream-VL by setting the training action chunk size to be larger than 12 and varying the predicted action chunk during inference time to see the performance of the two models in long-horizon action chunk prediction, as shown in Figure~\ref{fig:low_level}.

\begin{table}[h]
    \centering
    \caption{Performance comparison between autoregressive and diffusion-based VLMs on LIBERO benchmarks. Despite Qwen2.5-VL achieving stronger results on standard vision-language tasks (Table~\ref{tab:vl_sota_comparison1}-\ref{tab:video_results}), Dream-VL substantially outperforms it on robotic planning tasks, suggesting that diffusion-based modeling offers advantages for long-horizon action prediction.}
    \label{tab:libero_long_ablation}
    \resizebox{1\textwidth}{!}{%
    \begin{tabular}{lccccc}
        \toprule
        \textbf{Model} & \textbf{VLM Type} & \textbf{Size} & \textbf{Robotics Pretraining}  & \textbf{LIBERO-Goal (\%)} & \textbf{LIBERO-Long (\%)} \\
        \midrule
        OpenVLA & AR  & 7B & \textbf{\cmark} & 79.2 & 53.7 \\
        Qwen2.5-VL & AR  & 7B & \textbf{\xmark}  & 68.0 & 34.0 \\
        \cc \textbf{Dream-VL} & \cc \textbf{Diffusion} & \cc \textbf{7B} &\cc  \textbf{\xmark} & \cc \textbf{83.2} &\cc  \textbf{59.0} \\
        \bottomrule
    \end{tabular}%
    }
\end{table}

\paragraph{Results and Analysis.}
We have the following observations:
\begin{itemize}
\item 
As shown in Table \ref{tab:libero_long_ablation}, we observed a significant performance disparity between architectures. Although Qwen2.5-VL generally exhibits stronger performance on standard vision-language benchmarks in Table~\ref{tab:vl_sota_comparison1}, after SFT, it achieves only a 34.0\% success rate on LIBERO-Long. In contrast, the diffusion-based Dream-VL achieves 59.0\%, which also outperforms the large-scale robotic-pretrained OpenVLA. This result suggests that the diffusion-based VLM holds superior potential for learning long-horizon planning compared to autoregressive backbones.
\item As shown in Figure~\ref{fig:low_level}, Qwen2.5-VL tends to accumulate errors more easily when generating too many actions at once. This can be seen from the fact that Qwen2.5-VL reaches the maximum benefit of action chunking faster than Dream-VL. For instance, the optimal chunk sizes for Qwen2.5-VL are 3 for LIBERO-Goal and 5 for LIBERO-Long, compared to 9 and 10 for Dream-VL. After reaching the maximum, Qwen2.5-VL is affected by error accumulation from autoregressive generation. When this error accumulation becomes excessive, action chunking can even have a negative impact. For example, predicting and executing 9 or more actions at once in the LIBERO-Goal task performs worse than predicting and executing just one action at a time.
\item Dream-VL can maximize its parallel generation capability when predicting low-level actions. We found that for predicting multiple low-level actions, only one diffusion step is needed to achieve good performance, resulting in a significant speed advantage. This is different from text generation, where predicting multiple tokens in one step is often insufficient.
As shown in the figure on the right, generating 12 actions at once can achieve a 27$\times$ speedup. 
\end{itemize}

\begin{mybox}[colback=gray!10]{Highlights}
    \begin{itemize}
      \item Unlike autoregressive models (e.g., Qwen2.5-VL) where performance degrades with larger chunk sizes due to error accumulation, dVLMs such as Dream-VL maintain robustness over longer horizons. 
      \item Dream-VL exhibits exceptional efficiency in predicting low-level actions; remarkably, it requires only a single diffusion step to achieve competitive performance, delivering a 27$\times$ speedup over autoregressive generation.
    \end{itemize}
\end{mybox}

\section{Dream-VLA: Dream-VL with Large-scale Robotic Pretraining}

Given the potential of Dream-VL in low-level action planning, we further conduct a large-scale robotic pretraining over Dream-VL to obtain a general vision-language-action (VLA) model.

\begin{figure}[h]
    \centering
    \includegraphics[width=0.9\textwidth]{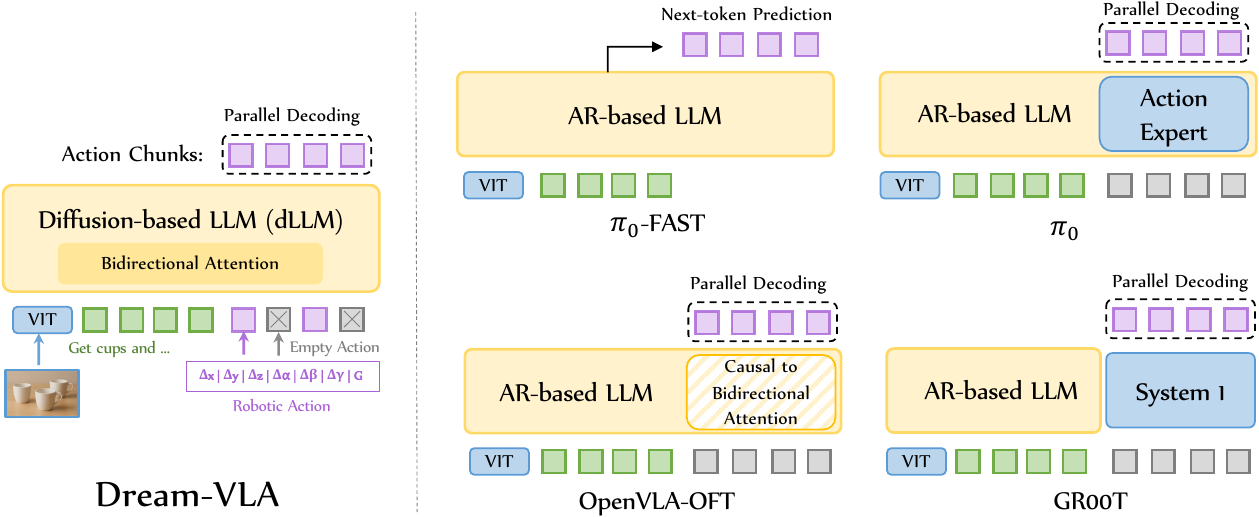}
    \caption{Architectural comparison of various VLA models.
    }
    \label{fig:comparison}
\end{figure}

\subsection{Setup.}

\paragraph{Robotic Pretraining.}
Following OpenVLA~\citep{openvla}, we use a large, diverse dataset of 970k robot manipulation trajectories from the Open-X Embodiment dataset~\citep{oxe}, which spans a wide range of robot embodiments, tasks, and scenes. Similar to training Dream-VL, we also use the same discrete diffusion loss as in the backbone language model Dream 7B during robotic pretraining. We continue training Dream-VL with a global batch size of 1024, a constant learning rate of 1e-5, an action chunk size of 8, and a total of 610,000 training steps.

\paragraph{Evaluation.}
To assess Dream-VLA's capabilities, we conduct experiments across three diverse robotic manipulation benchmarks: 
\begin{itemize}
    \item \textbf{LIBERO}~\citep{liu2023libero}, which consists of four task suites designed
for studying lifelong learning in robotic manipulation: LIBERO-Spatial (same objects, different spatial layouts), LIBERO-Object (fixed layout, different objects), LIBERO-Goal (fixed objects and layout, different goals), and LIBERO-Long (longer tasks that span multiple objects, layouts, and operations). LIBERO is built on a Franka arm with diverse scenes and expert demonstrations for each task; 
\item  \textbf{SimplerEnv}~\citep{simplerenv}, a simulated evaluation suite for real robot setups, including Google Robot tasks with Visual Matching and Variant Aggregation metrics across diverse visual variations, and WidowX robot tasks aligned with BridgeData-V2~\citep{walke2023bridgedata}; 
\item \textbf{Real-World PiPER Robot}, a tabletop manipulation evaluation on our PiPER platform with a single front-facing camera observing the robot and the table workspace. Following our concurrent work, we finetune only on simulation data synthesized with domain randomization (e.g., textures, scene, lighting, and camera pose), and evaluate in two settings: (i) sim2sim evaluation in simulation on picking up varying target objects randomly placed on the table in the presence of multiple distractors without test-time domain randomization, and (ii) zero-shot sim2real transfer on the real PiPER robot without any real-world finetuning. 
\end{itemize}

\paragraph{Finetuning.}
Following standard finetuning protocols~\citep{openvlaoft}, our policy takes as input RGB observations—specifically, one third-person camera and one wrist-mounted camera for LIBERO, or a single third-person camera for SimplerEnv—along with natural language task descriptions. Proprioceptive end-effector positions are optionally provided when available. 
When finetuning for downstream tasks, we can use any finetuning objective without changing the model architecture (e.g., L1 regression, continuous diffusion, discrete in OpenVLA-OFT~\citep{openvlaoft}, and discrete diffusion~\citep{discretediffusion}). To predict as fine-grained a move as possible, we prioritize a continuous action approach. By default, we use a Flow Matching~\citep{lipman2023flow} loss similar to $\pi_0$~\citep{black2024pi0} but without introducing a separate action expert, and the flow matching timesteps are set to 4 during inference. For SimplerEnv–Fractal and SimplerEnv–Bridge, we fine-tune on Fractal~\citep{rt1} and BridgeData V2~\citep{walke2023bridgedata}, respectively. We use an action chunk size of 8 for LIBERO and 5 for SimplerEnv to match the widely used settings. The fine-tuning is launched with LoRA~\citep{hu2022lora} with rank 32, batch size 64, and learning rate 1e-4. We train Dream-VLA on LIBERO for 100k steps and 200k for SimplerEnv, without careful checkpoint selection.

\begin{table}[t]
    \centering
    \caption{Success rate (\%) on LIBERO benchmark suites. We compare Dream-VLA with state-of-the-art policies and VLA models. ``-'' indicates the metric is not reported in the original paper.}
    \label{tab:libero_results}
    \resizebox{\textwidth}{!}{%
    \begin{tabular}{lcccccc}
        \toprule
        \textbf{Model} & \textbf{VLM Type} & \textbf{Spatial} & \textbf{Object} & \textbf{Goal} & \textbf{Long} & \textbf{Average} \\
        \midrule
        Diffusion Policy~\citep{diffusion_policy} & - & 78.3 & 92.5 & 68.3 & 50.5 & 72.4 \\
        MDT~\citep{MDT} & - & 78.5 & 87.5 & 73.5 & 64.8 & 76.1 \\
        Seer (scratch)~\citep{seer} & - & - & - & - & 78.7 & - \\
        Octo-Base~\citep{octo} & - & 78.9 & 85.7 & 84.6 & 51.1 & 75.1 \\
        Seer (fine-tuned)~\citep{seer} & - & - & - & - & 87.7 & - \\
        DiT Policy~\citep{dita}  & - & 84.2 & 96.3 & 85.4 & 63.8 & 82.4 \\
        \midrule
        TraceVLA~\citep{tracevla}  & AR & 84.6 & 85.2 & 75.1 & 54.1 & 74.8 \\
        SpatialVLA~\citep{spatialvla} & AR & 88.2 & 89.9 & 78.6 & 55.5 & 78.1 \\
        $\pi_0$ + FAST~\citep{pifast}& AR & 96.4 & 96.8 & 88.6 & 60.2 & 85.5 \\
        $\pi_0$~\citep{black2024pi0} & AR & 96.8 & 98.8 & 95.8 & 85.2 & 94.2 \\
        OpenVLA~\citep{openvla} & AR & 84.7 & 88.4 & 79.2 & 53.7 & 76.5 \\
        CoT-VLA~\citep{zhao2025cotvla} & AR &87.5 &91.6 &87.6&69.0 &81.1 \\
        OpenVLA-OFT~\citep{openvlaoft} & AR & 97.6 & 98.4 & \textbf{97.9} & 94.5 & 97.1 \\
        GR00T-N1~\citep{gr00t} & AR & 94.4 & 97.6 & 93.0 & 90.6 & 93.9 \\
        Discrete Diffusion VLA~\citep{discretediffusion} & AR & 97.2 & 98.6 & 97.4 & 92.0 & 96.3 \\
        \textbf{\cc Dream-VLA (Ours)} & \textbf{ \cc  Diffusion} & \textbf{\cc 97.6} & \textbf{\cc 98.8} & {\cc 97.2} & \textbf{\cc 95.0} & \textbf{\cc 97.2} \\
        \bottomrule
    \end{tabular}%
    }
\end{table}

\begin{table*}[t]
\centering
\caption{Real-world performance evaluation on WidowX Robot tasks. We report the Grasping Success Rate (Grasp) and Task Success Rate (Succ) for each task, along with the average task success rate. All values are in percentages (\%).}
\label{tab:windowx}
\resizebox{\textwidth}{!}{%
\begin{tabular}{lcccccccccc}
\toprule
\multirow{2}{*}{\textbf{Method}} & \multicolumn{2}{c}{\textbf{Spoon on Towel}} & \multicolumn{2}{c}{\textbf{Carrot on Plate}} & \multicolumn{2}{c}{\textbf{Stack Green Block}} & \multicolumn{2}{c}{\textbf{Eggplant in Basket}} & \multicolumn{2}{c}{\textbf{Average}} \\
\cmidrule(lr){2-3} \cmidrule(lr){4-5} \cmidrule(lr){6-7} \cmidrule(lr){8-9} \cmidrule(lr){10-11}
 & Grasp & Success & Grasp & Success & Grasp & Success & Grasp & Success & Success & Overall \\
\midrule
Octo-Base~\citep{octo} & 50.0 & 33.0 & 50.0 & 25.0 & 29.2 & 0.0 & 40.0 & 23.3 & 20.3 & 31.3 \\
RoboVLM~\citep{robovlm} & 37.5 & 20.8 & 33.3 & 25.0 & 8.3 & 8.3 & 0.0 & 0.0 & 13.5 & 16.7 \\
SpatialVLA~\citep{spatialvla} & 20.8 & 16.7 & 29.2 & 25.0 & 62.5 & 29.2 & 100.0 & 100.0 & 42.7 & 47.9 \\
RT-1-X~\citep{o2024open} & 4.2 & 0.0 & 16.7 & 0.0 & 0.0 & 0.0 & 3.3 & 0.0 & 0.0 & 3.0 \\
OpenVLA~\citep{openvla} & 4.1 & 0.0 & 33.0 & 0.0 & 12.5 & 0.0 & 8.3 & 4.1 & 1.0 & 7.8 \\
OpenVLA-OFT~\citep{openvlaoft} & 50.0 & 12.5 & 41.7 & 4.2 & 70.8 & 20.8 & 91.7 & 37.5 & 18.8 & 41.2 \\
$\pi_0$~\citep{black2024pi0} & 45.8 & 29.1 & 25.0 & 0.0 & 50.0 & 16.6 & 91.6 & 62.5 & 27.1 & 40.1 \\
$\pi_0$+FAST~\citep{pifast} & 62.5 & 29.1 & 58.5 & 21.9 & 54.0 & 10.8 & 83.3 & 66.6 & 32.1 & 48.3 \\
GR00T-N1~\citep{gr00t} & 83.3 & 62.5 & 54.2 & 45.8 & 70.8 & 16.7 & 41.7 & 20.8 & 36.5 & 49.5 \\
DiscreteDiffusionVLA~\citep{discretediffusion} & 70.8 & 29.2 & 58.3 & 29.2 & 62.5 & 20.8 & 91.7 & 70.8 & 37.5 & 54.2 \\
LLaDA-VLA~\citep{lladavla} & - & 56.9 & - & \textbf{76.3} & - & \textbf{30.6} & - & 58.3 & 55.5 & -  \\
\textbf{\cc Dream-VLA (Ours)} & \textbf{\cc 91.7} & \textbf{\cc 79.2} & \textbf{\cc 58.3} & {\cc 41.7} & \textbf{\cc 79.2} & {\cc 20.8} & \textbf{\cc 100.0} & \textbf{\cc 100.0} & \textbf{\cc 60.4} & \textbf{\cc 71.4} \\
\bottomrule
\end{tabular}%
}
\end{table*}

\subsection{Main Results}
\paragraph{LIBERO results.} 
We evaluate Dream-VLA on the LIBERO benchmark, with results shown in Table~\ref{tab:libero_results}. Our model demonstrates strong performance across four task suites, reaching 97.6\% on LIBERO-Spatial, 98.8\% on LIBERO-Object, 97.2\% on LIBERO-Goal, and 95.0\% on LIBERO-Long, yielding an overall average of 97.2\%. Dream-VLA shows that a model pretrained on a diffusion backbone can achieve superior performance than previous state-of-the-art autoregressive VLA model (OpenVLA-OFT, 97.1\% average).
These results show that diffusion-based modeling offers significant potential for vision-language-action tasks, providing a strong alternative approach to VLA modeling.

\paragraph{WidowX Robot results.} 
Table~\ref{tab:windowx} presents real-world evaluation results on the WidowX robot across four manipulation tasks. Dream-VLA achieves an overall average performance of 71.4\%, establishing new state-of-the-art results among VLA models. Our approach substantially outperforms diffusion/flow-matching-based policies such as $\pi_0$ (40.1\%) by 31.3 percentage points. When compared to autoregressive baselines, Dream-VLA surpasses leading methods such as OpenVLA-OFT (33.3\%), demonstrating the effectiveness of diffusion-based modeling.
Among diffusion-based approaches, Dream-VLA outperforms DiscreteDiffusionVLA by 17.2 percentage points. Unlike DiscreteDiffusionVLA, which adapts diffusion modeling from pretrained autoregressive VLA model, Dream-VLA builds from dLLM (Dream-7B) through sufficient vision-language alignment followed by VLA pretraining, thereby fully unlocking the potential of dLLMs for robotic manipulation tasks. 
The per-task results reveal consistent gains across both grasp and task success metrics. Dream-VLA demonstrates particularly strong performance on tasks such as Put Eggplant (100\% task success) and Put Spoon on Towel (79.2\% task success), highlighting its capability in fine-grained manipulation.

\begin{table}[t]
    \centering
    \caption{Performance evaluation on Google Robot tasks. All values are success rates in percentages (\%). ``-'' denotes missing data.}
    \label{tab:google_robot_results}
    \resizebox{\textwidth}{!}{%
    \begin{tabular}{lccccccccc}
        \toprule
        \multirow{2}{*}{\textbf{Method}} & \multicolumn{4}{c}{\textbf{Visual Matching}} & \multicolumn{4}{c}{\textbf{Variant Aggregation}} & \\
        \cmidrule(lr){2-5} \cmidrule(lr){6-9}
         & \textbf{Pick Coke} & \textbf{Mv Near} & \textbf{Drawer} & \textbf{Avg} & \textbf{Pick Coke} & \textbf{Mv Near} & \textbf{Drawer} & \textbf{Avg} & \textbf{Overall} \\
        \midrule
        RT-1-X~\citep{o2024open} & 56.7 & 31.7 & 59.7 & 53.4 & 49.0 & 32.3 & 29.4 & 39.6 & 46.5 \\
        RT-2-X~\citep{o2024open} & 78.7 & 77.9 & 25.0 & 60.7 & 82.3 & 79.2 & 35.3 & 64.3 & 62.5 \\
        Octo-Base~\citep{octo} & 17.0 & 4.2 & 22.7 & 16.8 & 0.6 & 3.1 & 1.1 & 1.1 & 9.0 \\
        OpenVLA~\citep{openvla} & 16.3 & 46.2 & 35.6 & 27.7 & 54.5 & 47.7 & 17.7 & 39.8 & 33.8 \\
        HPT~\citep{HPT} & 56.0 & 60.0 & 24.0 & 46.0 & - & - & - & - & - \\
        Moto~\citep{moto} & 74.0 & 60.4 & 43.1 & 59.2 & - & - & - & - & - \\
        RoboVLM~\citep{robovlm} & 77.3 & 61.7 & 43.5 & 63.4 & 75.6 & 60.0 & 10.6 & 51.3 & 57.4 \\
        TraceVLA~\citep{tracevla} & 28.0 & 53.7 & 57.0 & 42.0 & 60.0 & 56.4 & 31.0 & 45.0 & 43.5 \\
        $\pi_0$~\citep{black2024pi0} & 72.7 & 65.3 & 38.3 & 58.8 & 75.2 & 63.7 & 25.6 & 54.8 & 56.8 \\
        $\pi_0$+FAST~\citep{pifast} & 75.3 & 67.5 & 42.9 & 61.9 & 77.6 & \textbf{68.2 }& \textbf{31.3 }& \textbf{59.0} & 60.5 \\
        OpenVLA-OFT~\citep{openvlaoft} & 72.3 & 69.6 & 47.2 & 63.0 & 65.3 & 59.0 & 12.2 & 45.5 & 54.3 \\
        GR00T-N1~\citep{gr00t}& 47.0 & 70.0 & 18.1 & 45.0 & 78.8 & 62.5 & 13.2 & 51.5 & 48.4 \\
        Discrete Diffusion VLA~\citep{discretediffusion} & \textbf{85.4} & 67.5 & \textbf{60.6} & \textbf{71.2} & \textbf{82.5} & 64.6 & 23.6 & {56.9} & \textbf{64.1} \\
        \textbf{\cc Dream-VLA (Ours)} & {\cc 80.3} & \textbf{\cc 78.3} & {\cc 40.7} & {\cc 66.5} & {\cc 70.5} & {\cc 66.3} & {\cc 27.0} & {\cc 54.6} & {\cc 60.5} \\
        \bottomrule
    \end{tabular}%
    }
\end{table}

\paragraph{Google Robot results.}
Table~\ref{tab:google_robot_results} reports performance on Google Robot tasks in the SimplerEnv benchmark. Our model obtains an overall average of 60.5\%, performing on par with $\pi_0$+FAST and outperforming widely-adopted methods including $\pi_0$ (56.8\%), OpenVLA-OFT (54.3\%), and GR00T-N1 (48.4\%). Within the Visual Matching setting, Dream-VLA exhibits particularly strong results on Pick Coke (80.3\%) and Move Near (78.3\%) tasks. When examining the more challenging Variant Aggregation setting, Dream-VLA achieves performance (54.6\%) competitive with strong baseline $\pi_0$-FAST (59.0\%) and Discrete Diffusion VLA (56.9\%). These results show that Dream-VLA can generalize effectively across different visual conditions and tasks.

\subsection{Effectiveness of Robotic Pretraining}
\begin{figure}[h]
    \centering
    \includegraphics[width=0.9\textwidth]{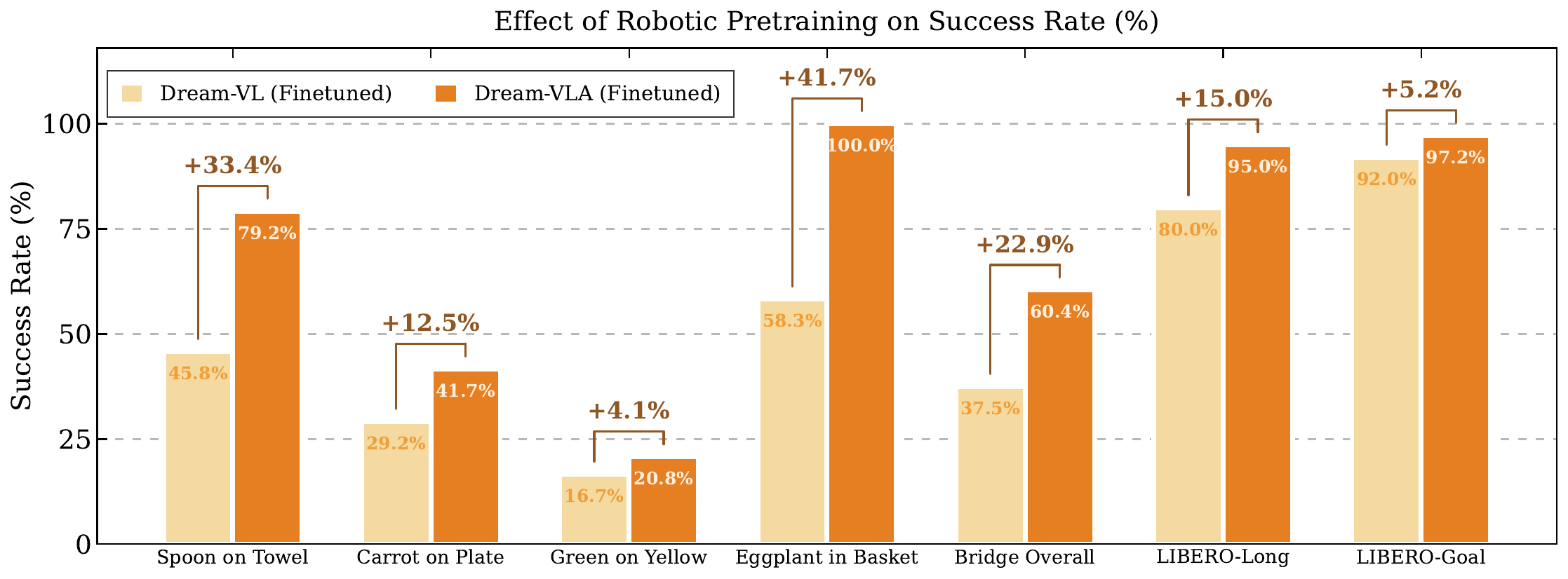}
    \caption{Performance gains from robotic pretraining across diverse tasks. Flow matching loss is used for all the downstream task fine-tuning.}
    \label{fig:vla_pretrain}
\end{figure}

Figure~\ref{fig:vla_pretrain} shows the impact of robotic pretraining. The result shows that robotic pretraining consistently improves fine-tuning outcomes across the majority of tasks, demonstrating its effectiveness for enhancing VLA capabilities. 
The magnitude of improvement varies across different tasks, with many showing substantial gains from pretraining, while a few tasks such as LIBERO-Goal and Stack Green Block On Yellow exhibit more modest improvements. The overall trend confirms that large-scale robotic pretraining provides valuable prior knowledge that transfers effectively to downstream tasks.

\subsection{What can we get from Dream-VLA?}
\paragraph{Architectural Consistency \& Superior Performance.}
For AR-based VLA backbones such as OpenVLA, downstream adaptation often requires structural modifications (i.e., attention mask adjustments) to enable action chunking, which is critical for AR-based models as it enhances both inference efficiency and task success rates. In contrast, DreamVLA, as a diffusion-based backbone, inherently supports action chunking without requiring architectural changes. In fact, the model architecture remains consistent starting from the LLM stage, thereby minimizing performance loss caused by structural variations across stages. As a result (Table~\ref{tab:all_sft}), we find that Dream-VLA consistently achieves superior downstream performance across various finetuning objectives.

\begin{table}[htbp]
\centering
\caption{Comparison of OpenVLA-OFT and Dream-VLA with different SFT training objectives on SimplerEnv-WidowX benchmark.}
\label{tab:all_sft}
\resizebox{\textwidth}{!}{
\begin{tabular}{lccccccccc}
\toprule
 & \multicolumn{2}{c}{\textbf{Spoon on Towel}} & \multicolumn{2}{c}{\textbf{Carrot on Plate}} & \multicolumn{2}{c}{\textbf{Stack Green Block}} & \multicolumn{2}{c}{\textbf{Eggplant in Basket}} & \textbf{Overall} \\
\cmidrule(lr){2-3} \cmidrule(lr){4-5} \cmidrule(lr){6-7} \cmidrule(lr){8-9} \cmidrule(lr){10-10}
\textbf{WidowX Robot} & Grasp & Success & Grasp & Success & Grasp & Success & Grasp & Success & Success \\
\midrule
\multicolumn{10}{l}{{\textbf{OpenVLA-OFT}}} \\
\hspace{3mm} w/ L1 & 62.5 & 20.8 & 45.8 & 20.8 & 66.7 & 16.7 & 100.0 & 87.5 & \cg 36.5 \\
\hspace{3mm} w/ Discrete & 91.7 & 29.2 & 58.3 & 12.5 & 58.3 & 8.3 & 95.8 & 75.0 & \cg 31.3 \\
\hspace{3mm} w/ Cont. Diffusion & 50.0 & 12.5 & 33.3 & 0.0 & 37.5 & 0.0 & 66.7 & 4.2 & \cg 4.2 \\
\hspace{3mm} w/ Disc. Diffusion & 75.0 & 37.5 & 58.3 & 37.5 & 58.3 & 12.5 & 87.5 & 50.0 & \cg 34.4 \\
\hspace{3mm} w/ Flow Matching & 58.3 & 12.5 & 45.8 & 4.2 & 37.5 & 0.0 & 91.7 & 25.0 & \cg 10.4 \\
\midrule
\multicolumn{10}{l}{{\textbf{Dream-VLA}}} \\
\hspace{3mm} w/ L1 & 75.0 & 62.5 & 66.7 & 54.2 & 79.2 & 8.3 & 100.0 & 100.0 & \cc 56.3 \\
\hspace{3mm} w/ Discrete & 70.8 & 37.5 & 62.5 & 41.7 & 79.2 & 8.3 & 95.8 & 91.7 & \cc 44.8 \\
\hspace{3mm} w/ Cont. Diffusion & 79.2 & 66.7 & 66.7 & 54.2 & 95.8 & 16.7 & 91.7 & 91.7 &\cc  57.3 \\
\hspace{3mm} w/ Disc. Diffusion & 79.2 & 45.8 & 62.5 & 45.8 & 83.3 & 25.0 & 100.0 & 87.5 & \cc 51.0 \\
\hspace{3mm} w/ Flow Matching & {91.7} & {79.2} & 58.3 & 41.7 & 79.2 & 20.8 & {100.0} & {100.0} & \cc {60.4} \\
\bottomrule
\end{tabular}
}
\end{table}

\paragraph{Accelerated Convergence in Fine-tuning.}
Due to the absence of architectural modifications, we find the model tends to converge faster and achieves a lower loss in finetuning as compared to OpenVLA-OFT. The difference is most pronounced under discrete diffusion finetuning because it shares the same training objective with LLM, VLM, and VLA pretraining.

\begin{figure}[h]
    \centering
    \includegraphics[width=1\textwidth]{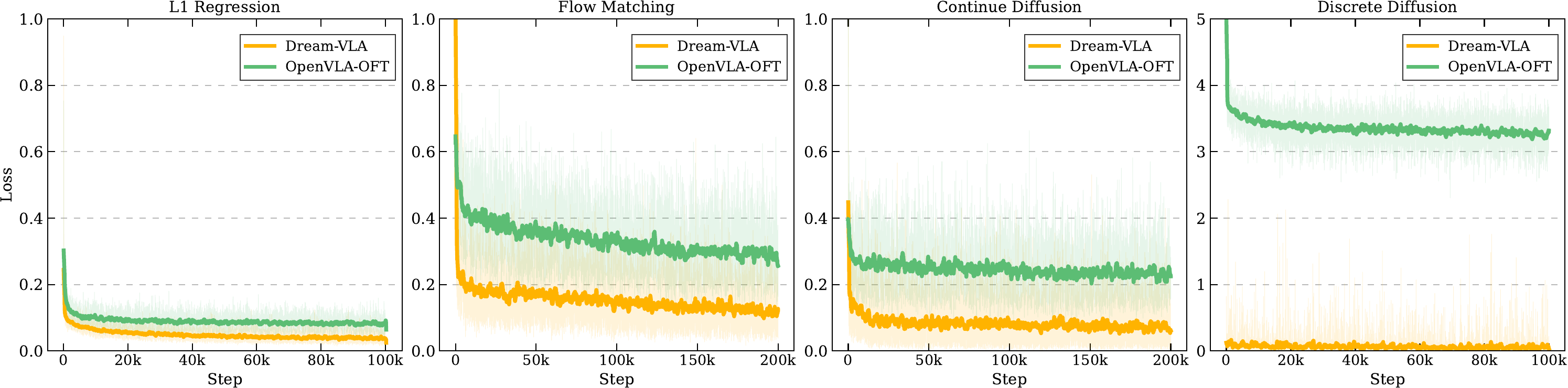}
    \caption{Loss curves of Dream-VLA and OpenVLA-OFT under different training objectives.}
    \label{fig:loss_curve}
\end{figure}

\begin{mybox}[colback=gray!10]{Highlights}
    \begin{itemize}
        \item Dream-VLA achieves a new state-of-the-art on the WidowX Robot with 71.4\% overall average score, substantially outperforming all prior methods (previous best 54.2\%), and achieves a new state-of-the-art result on the LIBERO (97.2\%).
      \item Dream-VLA consistently outperforms the strong AR-based baseline OpenVLA-OFT across diverse finetuning objectives, demonstrating the advantage of a diffusion-based backbone.
      \item Diffusion-based backbone requires no architectural modifications compared to AR-based VLA models to enable action chunking, providing inherent advantages in structural consistency and faster training convergence.
    \end{itemize}
\end{mybox}

\subsection{Real-World Performance}

\paragraph{PiPER Sim2Sim evaluation.}
We first benchmark Dream-VLA under the PiPER sim-to-sim tabletop pick setting and compare against a strong autoregressive baseline, OpenVLA-OFT. We finetune both models on domain-randomized simulation data with L1 regression using the same LoRA setting as mentioned above, for 60k steps. We use an action chunk size of 50 and execute the first 20 steps for each action chunk. Two evaluation protocols are considered: (i) a simplified setting where both training and evaluation are restricted to 10 objects; (ii) a full setting with 75 target objects. Each object is evaluated for 10 trials, where the target is randomly placed among several distractors.  Table~\ref{tab:piper_result} reports success counts for a representative subset of objects, 
while overall success rates are calculated over 100 and 750 total trials, respectively.
Overall, on this relatively simple pick-up setting, Dream-VLA slightly outperforms OpenVLA-OFT, suggesting that diffusion-based VLA modeling remains competitive under controlled sim2sim evaluation. In addition, Dream-VLA achieves a $\sim$1.31$\times$ training speedup over OpenVLA-OFT for the same number of steps, and the absolute wall-clock time savings grow with longer training schedules. In future work, we will extend the comparison to more diverse and challenging scenarios to better investigate the strengths of different models.

\begin{table}[htbp]
\centering
\caption{Comparison of OpenVLA-OFT and Dream-VLA under PiPER sim-to-sim. We report successes out of 10 trials for each object (for the 75-obj setting, only the first 10 objects are shown for readability). Overall is computed over all objects in the corresponding setting.
}
\label{tab:piper_result}
\resizebox{\textwidth}{!}{
\small
\begin{tabular}{llcccccccccccc}
\toprule
    Setting & Model     & Apple & Avocado & Banana & Bear & Melon & Bowl & Tea   & Brush & Cabbage & Camel & Others & Overall  \\
\midrule
\multirow{2}{*}{\textbf{10 objs}}
& \textbf{OpenVLA-OFT} & 10/10 & 9/10     & 10/10 & 10/10 & 10/10 & 5/10 & 9/10  & 10/10 & 10/10 & 10/10 & -- & \cg 93.00\%  \\
& \textbf{Dream-VLA}   & 10/10 & 10/10    & 10/10 & 10/10 & 10/10 & 5/10 & 10/10 & 9/10  & 10/10 & 10/10 & -- & \cc 94.00\%  \\
\midrule
\multirow{2}{*}{\textbf{75 objs}}
& \textbf{OpenVLA-OFT} & 6/10  & 6/10     & 9/10  & 10/10  & 9/10  & 5/10 & 8/10 & 10/10 & 9/10  & 10/10 & \multicolumn{1}{c}{$\cdots$} & \cg 81.87\%  \\
& \textbf{Dream-VLA}   & 10/10 &10/10     & 9/10  & 10/10  & 10/10 & 3/10 & 9/10 & 10/10 & 10/10 & 10/10 & \multicolumn{1}{c}{$\cdots$} & \cc 83.73\% \\
\bottomrule
\end{tabular}
}
\end{table}

\paragraph{PiPER Sim2Real evaluation.}
We further perform zero-shot sim-to-real transfer on the real PiPER robot picking task using a single front-facing Intel RealSense D455 camera, without any real-world fine-tuning. Since the camera pose is randomized during training, we do not require rigorous extrinsic calibration at test time; instead, we place the camera in front of the workspace and manually ensure that its view roughly matches the simulated viewpoint. Nevertheless, we observe that camera placement can still have a noticeable impact on real-world performance. Due to unavoidable variations in scene resets, lighting, and camera pose across trials, we do not aim for a strictly controlled quantitative comparison. Qualitatively, Dream-VLA can reliably pick up a wide range of objects under moderate appearance shifts, demonstrating strong generalization. Remaining failures mainly stem from residual sim-to-real gaps (e.g., viewpoint and appearance discrepancies), suggesting that better calibration and improved robustness to visual shifts would further benefit real-world deployment.
\section{Conclusion and Future Work}
\label{sec:conclusion}

In this paper, we release Dream-VL and Dream-VLA as preliminary attempts at dLLMs for visual planning. Dream-VL achieves competitive performance with current open-source state-of-the-art AR VLMs on general visual understanding tasks and significantly surpasses previous dVLMs. Although general capabilities still lag behind top closed-data AR models, Dream-VL demonstrates significant advantages for tasks requiring global planning. Followed with robotic pretraining, Dream-VLA stands as a pioneering pretrained VLA based on dLLMs, achieving strong performance and showcasing its considerable promise.

However, there is still much room for improvement. Firstly, we do not systematically study data in this work; both the VL and VLA training data largely follow prior setups. There is still substantial room on this side to further improve performance. Second, an interesting direction is mixture training that jointly covers high-level planning and low-level control, such as in RT-2~\citep{rt2}. Since dLLMs already exhibit advantages over AR LLMs in high-level planning, it is promising to investigate VLA backbones based on dLLMs that jointly improve both high-level plan generation and low-level action prediction.
Thirdly, we empirically observe that continuous action spaces often outperform discrete ones in many downstream SFT settings. This suggests that (i) continuous-space robotic pretraining over discrete dVLMs or continuous dVLMs is a promising direction, and (ii) on top of a discrete dVLM, designing better discrete action representations such as FAST~\citep{pifast} could further close the gap for discrete VLAs.
Finally, our current real-robot experiments are still preliminary. A natural next step is to collect larger and more diverse real-world datasets to systematically evaluate how Dream-VLA behaves in realistic environments, and how it performs compared with AR-VLM-based VLAs regarding generalization.

\section*{Acknowledgments}
We thank Zhili Liu, Lei Li, Yiheng Xu, and Mukai Li for the valuable discussion on VLM, and Shengliang Deng for the valuable discussion on VLA.

\bibliographystyle{plainnat}
\bibliography{ref}

\end{document}